# LARSEN-ELM: Selective Ensemble of Extreme Learning Machines using LARS for Blended Data

Bo Han[a], Bo He[a,*], Rui Nian[a], Mengmeng Ma[a], Shujing Zhang[a], Minghui Li[b], Amaury Lendasse[c]

[a]*School of Information and Engineering, Ocean University of China, Shandong, Qingdao, China 266000*
[b]*Centre for Ultrasonic Engineering, University of Strathclyde, Royal College Building, 204 George Street, Glasgow G1 1XW, United Kingdom*
[c]*Department of Information and Computer Science, Aalto University, Espoo, Finland 00076*

**Abstract**

Extreme learning machine (ELM) as a neural network algorithm has shown its good performance, such as fast speed, simple structure etc, but also, weak robustness is an unavoidable defect in original ELM for blended data. We present a new machine learning framework called "LARSEN-ELM" for overcoming this problem. In our paper, we would like to show two key steps in LARSEN-ELM. In the first step, preprocessing, we select the input variables highly related to the output using least angle regression (LARS). In the second step, training, we employ Genetic Algorithm (GA) based selective ensemble and original ELM. In the experiments, we apply a sum of two sines and four datasets from UCI repository to verify the robustness of our approach. The experimental results show that compared with original ELM and other methods such as OP-ELM, GASEN-ELM and LSBoost, LARSEN-ELM significantly improve robustness performance while keeping a relatively high speed.

## 1. Introduction

One reason why feed-forward neural networks are rarely used in the industry or real-time area is that networks need large amounts of time and training data to perform well. When looking into the question carefully, we had to face two challenging issues over the past few decades: first, why was the computation of neural network so slow? And second, how to solve the bottleneck in the applications [1]? The primary reasons may be: firstly, many parameters need to be tuned manually in several layers [2,3]. Secondly, it needs to repeat learning several times in order to form a good mode like BP-algorithm [4,5].

Recently, Guang-Bin Huang et al. proposed a novel learning algorithm called extreme learning machine (ELM) which has a faster speed and higher generalization performance [6,7]. As an emerging algorithm, ELM was originally evolved from single hidden layer feed-forward networks (SLFN), which was extended to the generalized one by Huang et al. later. The essence of ELM can be concluded into two main aspects: on the one hand, the input weights and the hidden layer biases of the SLFN can be assigned randomly, which need not to be tuned manually. On the other hand, the output weights of the SLFN are computed by the generalized inverse of the hidden layer output matrix because it is simply treated as a linear network [1]. For the performance of ELM, it tends to reach not only the smallest training error but also the smallest norm of output weights. Huang et al. have investigated the interpolation capability and universal approximation capability of ELM, then completely studied the kernel based ELM which can be applied in regression, binary and multi-label classification tasks. For equalization problems, ELM based complex-valued neural networks using fully complex activation function have attracted considerable attention [8]. What's more, many other forms of ELM also develop very fast. The online sequential ELM is a simple but efficient learning algorithm which handles both additive and RBF nodes in the unified framework [9]. Although the incremental ELM (I-ELM) has no parameters for users to set except the maximum network architecture and the expected accuracy, I-ELM outperforms other learning algorithms, including support vector regression, stochastic gradient-descent BP in generalization performance and learning speed as well [10]. Frénay and Verleysen studied the kernel implementation of ELM in the complex space. Their contribution has shown that ELM could work for the conventional SVM and its variants, in addition, it could achieve better generalization [11]. Besides, Rong et al. presented a method called P-ELM as ELM classifier network [12].

---

\* Corresponding author
  *Email address*: bhe@ouc.edu.cn (Bo He).

Although, ELM has shown its good performance in real applications, such as fast speed, simple structure etc. However, many researchers pay more attention on the robustness of ELM caused by irrelevant input variables nowadays. Miche and Lendasse et al. proposed a pruning extreme learning machine called OP-ELM (Optimal-Pruned ELM) [13,14], improving the robustness of ELM and achieving greater accuracy due to its variables selection that removes the possibly irrelevant variables from blended data [14,15]. However, the OP-ELM does not work very well in high dimensional problems. Thus, we investigate the idea that a neural network ensemble may enhance the robustness of ELM, because Hansen and Salamon have shown that the performance of a single network can be improved using an ensemble of neural networks with a plurality consensus scheme [16]. The ELM ensemble was proposed by Sun et al. [17]. The resulting ensemble has a better generalization performance, and significantly improves the robustness of ELM while keeping high speed. However, if the raw data is blended with irrelevant input variables, for example gaussian noise [18,19], the ELM ensemble using the technique of weighted average does not work well any more. Under the assumption of independent ensemble components, Zhou at al. suggested that ensemble several models may be better than ensemble all of them, and a selective neural network ensemble based on genetic algorithm (GASEN) was proposed later. The selective ensemble utilizes fewer but better individual models to ensemble, which achieves stronger generalization ability. However, the GASEN is much slower than other ensemble algorithms in computation because it employs genetic algorithm to select an optimum set of individual networks.

Inspired by these observation, for blended data [20], we hope to create a new method, which not only keeps high speed, but also improves robustness performance. So a new approach called LARSEN-ELM is proposed in this paper. Our approach consists of two stages: First we employ least angle regression (LARS) to select the targeted inputs highly related to the outputs [21]. Then, we train several independent ELM models and select an optimal set using genetic algorithm to constitute an ensemble [22,23]. We choose a sum of two sines and four datasets from UCI repository [24] to verify the new method and compare it with ELM and other popular algorithms such as OP-ELM, GASEN-BP(GASEN), GASEN-ELM and Least Squares Boosting (LSBoost). In our experiments, the new method turns out to be more robust than ELM while keeping relatively high speed. However, the robustness performance is improved at the cost of space because selective ensemble initially needs amounts of independent ELMs.

The rest of this paper is organized as follows. In Section 2, we present basic knowledge about ELM, In Section 3, LARS selection, Genetic Algorithm based Selective Ensemble and our new method (LASEN-ELM) are analyzed individually. In Section 4, several experiments on ELM. OP-ELM, GASEN-BP, GASEN-ELM, LARSEN-ELM, and a kind of boosting algorithm called Least Squares Boosting(LSBoost) are reported. In Section 5, we present our discussions which come from the performance of the new method. Finally in Section 6, conclusions on the new research direction are presented and several issues for future work are indicated.

## 2. Extreme Learning Machine

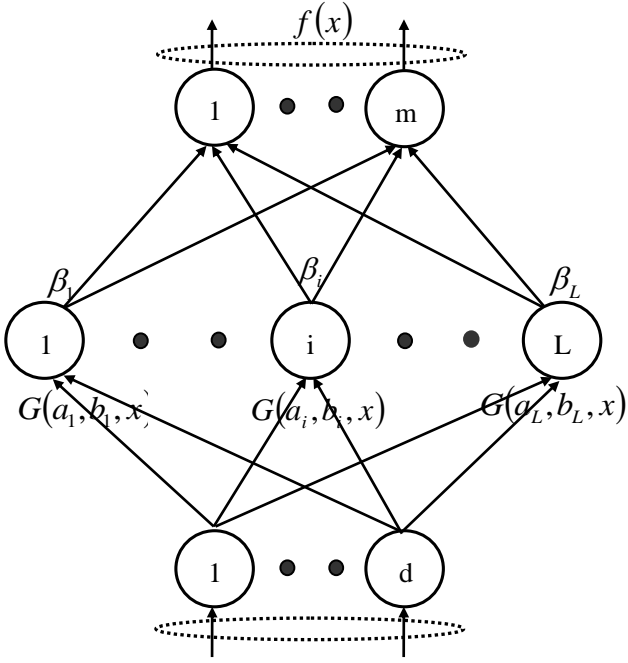

Figure 1. The structure of ELM

Guang-Bin Huang et al. raised a neural network algorithm called Extreme Learning Machine (ELM) [25,26], which takes full advantage of single-hidden layer feed-forward networks(SLFNs) to reduce the time of computation. It performs at fast learning speed and has a high generalization, both in regression problem and pattern recognition. Figure 1 reveals the structure of ELM. From the figure, we obviously note that ELM consists of one input layer, one hidden layer, and one output layer. The key to this structure is that the hidden layer needs not to be tuned iteratively [6,26], that is, the parameters of the hidden nodes which include input weights and biases can be randomly generated, so the problem boils down to the computation of output weights solved by the generalized inverse of the hidden layer output matrix. Compared with traditional learning algorithms, ELM has a simple structure and its learning speed is much faster than anything else. Hence, the ELM algorithm is popular among researchers [27-30].

For $N$ arbitrary distinct samples $(x_i, t_i) \in R^d \times R^m$, the model of SLFN with $L$ hidden nodes can be presented as:

$$\sum_{i=1}^{L} \beta_i G(a_i, b_i, x_j) = o_j, \quad j = 1, \ldots, N \quad (1)$$

Where $\beta_i = [\beta_{i1}, \beta_{i2}, \ldots, \beta_{im}]$ is the weight vector, connecting the $ith$ hidden node with the output nodes, $a_i$ is the weight of the $ith$ input node, $b_i$ is the threshold of the $ith$ hidden node, $x_j$ is the $jth$ input data, and $G(x)$ is the activation function of the SLFN. The outputs of SLFN can fully approximate the outputs of $N$ samples with zero error, that is, $\sum_{j=1}^{L} \|o_j - t_j\| = 0$, hence,

$$\sum_{i=1}^{L} \beta_i G(a_i, b_i, x_j) = t_j, \quad j = 1, \ldots, N \quad (2)$$

It can also be equivalent to the following equation:

$$H\beta = T \quad (3)$$

Where

$$H = \begin{bmatrix} G(a_1, b_1, x_1) \cdots G(a_L, b_L, x_1) \\ \vdots \\ G(a_1, b_1, x_N) \cdots G(a_L, b_L, x_N) \end{bmatrix}_{N \times L} \quad (4)$$

$$\beta = \begin{bmatrix} \beta_1^T \\ \vdots \\ \beta_L^T \end{bmatrix}_{L \times m} \quad \text{and} \quad T = \begin{bmatrix} t_1^T \\ \vdots \\ t_N^T \end{bmatrix}_{N \times m} \quad (5)$$

$H$ [12,27] is addressed as the hidden layer output matrix of the SLFNs. Hence, when the parameters $(a_i, b_i)$ of the hidden layer of SLFNs are randomly generated, $H$ matrix can be obtained given the training samples available, so the output weights $\beta$ can be obtained as follows:

$$\beta = H^+ T \quad (6)$$

Where $H^+$ is the Moore-Penrose generalized inverse of $H$ matrix [28,29]. So, after randomly generated, the hidden nodes parameters $(a_i, b_i)$ remain fixed. The output weights matrix $\beta$ can be obtained easily by the least-square method [30,31]. The algorithm ELM can be as follows:

**Algorithm 1. ELM:** Given a training set $N = \{(x_i, t_i) | x_i \in R^d, t_i \in R^m, i = 1, \ldots, N\}$, hidden node output function $G(a_i, b_i, x)$, and hidden node number $L$,
*Step* 1: Randomly generate input weights $a_i$ and biases $b_i$, $i = 1, \ldots, L$.
*Step* 2: Calculate the hidden layer output matrix $H$.
*Step* 3: Calculate the output weight vector $\beta$:

$\beta = H^+ T$.

## 3. Selective Ensemble of ELM using LARS for Blended Data

### 3.1 LARS Selection

Least Angle Regression (LARS) is introduced to handle variables selection for regression problems [31], which provides us the ranking of possible input variables. It is also a special case of Multiresponse Sparse Regression (MRSR) [21]. So based on MRSR[17,18], the first step of LARS selection ranks input variables by the degree related to the outputs, and the second step addresses the problem that how many input variables could be selected by the least mean square error between the prediction and expectation.

Suppose that the targets are indicated by matrix $T = [t_1 .. t_p]_{n \times p}$, regressors are indicated by matrix $X = [x_1 .. x_m]_{n \times m}$. MRSR adds sequentially active regressors to the model so that the $n \times p$ matrix $Y^k = [y_1^k .. y_p^k]$ models the targets $T$ appropriately as equation (7) shows.

$Y^k = XW^k$ (7)

The $m \times p$ weight matrix $W^k$ includes $k$ nonzero rows at the beginning of the $k$th step. LARS is MRSR in the situation of $p = 1$, that is, MRSR is an extension of LARS.

Set $k = 0$, initialize all elements of $Y^0$ and $W^0$ to zero, and normalize both $T$ and $X$ to zero mean. The scales of the columns of $T$ and the columns of $X$ should be equal, which may differ between the matrices. A cumulative correlation between the $j$th regressor $x_j$ and the current residuals is defined as:

$$c_j^k = \left\|(T - Y^k)^T x_j\right\|_1 = \sum_{i=1}^{p} \left|(t_i - y_i^k)^T x_j\right| \quad (8)$$

The formula measures the sum of absolute correlation between the residuals and regressors over all $p$ target variables at the beginning of the $k$th step. Let $c_{\max}^k$ indicate the maximum cumulative correlation, and A indicates the maximum among the group of regressors.

$$c_{\max}^k = \max_j \{c_j^k\}, A = \{j | c_j^k = c_{\max}^k\} \quad (9)$$

Let $n \times |A|$ matrix $X_A = [\ldots x_j \ldots]_{j \in A}$ be used to collect regressors which belong to A, so OLS (Ordinary Least Squares) estimation can be calculated by equation (10), and OLS estimation involves $k+1$ regressors at the $k$th step.

$$\bar{Y}^{k+1} = X_A (X_A^T X_A)^{-1} X_A^T T \quad (10)$$

A less greedy algorithm is defined by moving the MRSR estimation $Y^k$ toward the OLS estimation $\bar{Y}^{k+1}$, In contrast, the greedy forward selection adds regressors based on equation (8) and uses the OLS estimate (10). For example, in the direction $U^k = \bar{Y}^{k+1} - Y^k$, we will not achieve it. The major step is potentially taken in the direction of $U^k$ until some $x_j$, where $j \notin A$, has large cumulative correlation with the current residuals. The LARS estimate $Y^k$ is updated:

$$Y^{k+1} = Y^k + \gamma^k(\overline{Y}^{k+1} - Y^k) \tag{11}$$

To make the update, we should calculate the correct step size $\gamma^k$ first. According to equation (10), we know:

$$X_A^T(\overline{Y}^{k+1} - Y^k) = X_A^T(T - Y^k) \tag{12}$$

So:

$$c_j^{k+1}(\gamma) = |1-\gamma|c_{\max}^k \text{ for all } j \in A \tag{13}$$

$$c_j^{k+1}(\gamma) = \|a_j^k - \gamma b_j^k\|_1 \text{ for all } j \notin A \tag{14}$$

Here, $a_j^k = (T-Y^k)^T x_j$ and $b_j^k = (\overline{Y}^{k+1} - Y^k)^T x_j$. When equation (13) and equation (14) are equal, a new regressor with index $j \notin A$ will come into the model. So this happens only if the step size is taken from the set $\Gamma_j$ which satisfies equation (15) as follows:

$$\Gamma_j = \left\{ \frac{c_{\max}^k + s^T a_j^k}{c_{\max}^k + s^T b_j^k} \right\}_{s \in S} \tag{15}$$

$S$ is the set of $2^p$ sign vectors which is $p \times 1$ size, and the elements of $s$ may be either 1 or $-1$. The correct choice is the smallest size among such positive steps, which updates equation (11).

$$\gamma^k = \min\{\gamma | \gamma \geq 0 \text{ and } \gamma \in \Gamma_j \text{ for some } j \notin A\} \tag{16}$$

The weight matrix may be updated as follow:

$$W^{k+1} = (1-\gamma^k)W^k + \gamma^k \overline{W}^{k+1} \tag{17}$$

$\overline{W}^{k+1}$ is a sparse matrix of $m \times p$. So the parameters of selected regressors are shrunk by the equation (17).

**Algorithm 2. MRSR** (LARS is MRSR in the situation of $p = 1$)**:** Given a training set $\{(x_i, t_i) | x_i \in R^m, t_i \in R^p, i = 1,...,N\}$, for $k = 0,...,m$,

*Step* 1: Calculate cumulative correlation:

$$c_j^k = \sum_{i=1}^p \left|(t - y^k)^T x_j\right|$$

*Step* 2: Update the MRSR estimate:
$$Y^{k+1} = Y^k + \gamma^k(\overline{Y}^{k+1} - Y^k)$$
*Step* 3: Update weight matrix:
$$W^{k+1} = (1-\gamma^k)W^k + \gamma^k \overline{W}^{k+1}$$
Step 4: Get the ranking of input variables.

*3.2 Genetic Algorithm based Selective Ensemble*

*3.2.1 Selective Ensemble*

Neural network ensemble is a learning paradigm where several networks are jointly used to solve problems. But it is not perfect just ensembling all individual neural networks. when the relationship between the generalization of ensemble network and the correlation of individual network is analyzed, Zhi-Hua Zhou et al. proposed a new method called selective ensemble. Firstly, we assign each network a weight which reflects its importance in set, then after evolutionary learning, we can ensemble a set of optimal networks whose weight is bigger than the threshold predefined previously.

*3.2.2 Genetic Algorithm*

The Genetic Algorithm (GA) is designed to simulate evolutionary processes in the natural system, so the essence of GA is to search for a pool of candidate hypotheses and determine the best hypothesis. In GA, if the fitness is predefined as a numerical measure for specific problems, then a hypothesis leading to optimal fitness is defined as the "best hypothesis". Although many implementations of GA are different in details, they all share a common structure as follows: The algorithm iteratively updates a pool of hypothesis called population. In each iteration, all members from the population are evaluated by the fitness function. Then we select the fittest individuals from current population to produce a new one. Some of selected individuals are directly added into the next generation population. Others are viewed as sources for generating new offspring individuals according to genetic methods such as crossover and mutation [33].

**Algorithm 3. GA** ($Fitness, Fitness_t, p, r, m$)

$Fitness$: A function that assigns an evaluation score, given a hypothesis.

$Fitness_t$: A threshold specifying the termination criterion.

$p$: The number of hypotheses to be included in the population.

$r$: The fraction of the population to be replaced by Crossover at each step.

$m$: The mutation rate.

Step 1: $P \leftarrow$ initialize population: generate $p$ hypotheses at random.

Step 2: Evaluate: for each hypothesis $h_i$ in $P$, compute $Fitness(h_i)$.

Step 3: while $\max_h Fitness(h_i) < Fitness_t$, do

Create a new generation, denoted by $P_S$:

1. Select: probabilistically select $(1-r)p$ members of $P$ to be included in $P_S$. The probability $\Pr(h_i)$ of selecting hypothesis $h_i$ from $P$ is given by

$$\Pr(h_i) = \frac{Fitness(h_i)}{\sum_{j=1}^{p} Fitness(h_i)} \quad (18)$$

2. Crossover: probabilistically select $r \cdot p$ pairs of hypotheses from $P_S$, according to $\Pr(h_i)$ given above. For each pair $[h_1, h_2]$, produce two offspring by applying the Crossover operator. Add all offspring to $P_S$.

3. Mutate: choose $m$ percent of the members of $P_S$, with uniform probability. For each, flip one randomly-selected bit in its representation.

4. Update: $P \leftarrow P_S$.

5. Evaluate: for each hypothesis $h_i$ in $P$, compute $Fitness(h_i)$.

Step 4: Return the hypothesis $h_i$ from $P$ that has the highest fitness.

*3.2.3 GASEN*

In selective ensemble, assuming that weight of the $ith$ individual network is $\omega_i$, then we get a weight vector $w = [\omega_1, \omega_2, ..., \omega_N]$, and $\omega_i$ satisfy equation (19) and (20).

$$0 \leq \omega_i \leq 1 \quad (19)$$

$$\sum_{i=1}^{N} \omega_i = 1 \quad (20)$$

In GASEN, we employ genetic algorithm to evolve an optimal weight vector $w_{opt}$ from a pool of different weight vector $w$ while selecting components of $w_{opt}$ by predefined threshold $\lambda$. But it is notable that the sum of components of weight vector $w$ no longer satisfies equation (19) during evolution. So, only after we normalize the optimal weight vector, its components can be compared with predefined threshold. And the normalization of selective ensemble as follows:

$$\omega_i' = \omega_i / \sum_{i=1}^{N} \omega_i \quad (21)$$

Then, the output of selective ensemble is the average of those selected individual neural networks. The selective ensemble estimates a generalization error using a validation set. We use $\hat{E}_w^V$ to express the goodness of $w$. Then obviously, the smaller $\hat{E}_w^V$ is, the better $w$ is. So, the selective ensemble uses $f(w) = 1/\hat{E}_w^V$ as the fitness function during evolution. Finally, we can select the individuals with larger weight to compose an optimal ensemble.

The validation set can be generated in a variety of ways such as using an independent validation set, or generating repeat sampling from the training set. The selective ensemble uses a common approach to generate individual neural networks like bagging [32]. Supposing the whole original training set is $S$, and $S_i$ is the validation set of individual neural network $f_i$ generated by repeatable sampling from $S$. $S_i$ and $S$ are almost similar scale, then the data included in $S$ but not in $S_i$ is indicated in equation (22).

$$(1-\frac{1}{N})^N |S| \approx 0.368|S| \qquad (22)$$

Equation (21) shows that there is about 1/3 data in the training set $S$ not appearing in the validation set $S_i$. Let us assume that $\bar{S}_i$ consists of these data, so $\bar{S}_i$ is considered as the validation set of $f_i$ to assist training. In addition, since each network only uses a portion of data in $S$, so it can be used as the validation set during evolution. Of course, this can eliminate the extravagant demand for an additional validation set.

*3.3 LARSEN-ELM*

Inspired by the observation to drawbacks of ELM, we hope to find an ensemble algorithm to improve the robustness performance of ELM because many could be better than one [34]. Zhao et al. proposed an algorithm called GASEN-ELM to enhance the effluent quality predictions [35,36]. However, the method does not work well for blended data because irrelevant variables still disturb the robustness performance of network.

We present a new approach called LARSEN-ELM in order to improve the robustness performance of ELM for blended data. It consists of two significant steps. In the step of preprocessing, we use LARS to select targeted inputs which are highly related to the outputs, then, in the step of training, we take advantage of selective ensemble to constitute a set of optimal network.

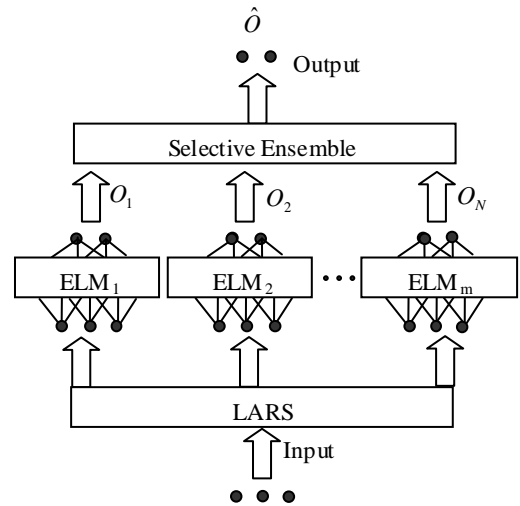

Figure 2. LARSEN-ELM

Just as the Figure 2, blended data, coming into the system, will be filtered by LARS for eliminating noise data as much as possible. This step can improve the reliability of the original data, and the principle of LARS is described in section 3.1. The selected data will be sent to the different ELM models to train, and finally get the output by selective ensemble. On behalf of generating the ensemble ELM with smaller population but stronger robustness, we can utilize genetic algorithm to select an optimal subset from a set of available ELMs as Zhou et al. did. We can sample $x$ according to a probability distribution $p(x)$ and the target output is $y$, and the output of the

$ith$ ELM is $f_i(x)$. Initially, we assign random weights vector $w = [\omega_1, \omega_2, ..., \omega_N]$ to the ensemble network where $\omega_i$ is the weight of the $ith$ individual network. Besides $\omega_i$ meets the requirement of equation (19) (20). If we complete the selection of ensemble ELM models, the output is:

$$\hat{f}(x) = \sum_{i=1}^{N} \omega_i f_i(x) \tag{23}$$

The error $E_i(x)$ of the $ith$ ELM on input $x$ and the error $\hat{E}(x)$ of the ensemble on input $x$ are respectively:

$$E_i(x) = (f_i(x) - y)^2 \tag{24}$$

$$\hat{E}(x) = (\hat{f}(x) - y)^2 \tag{25}$$

Then the generalization error $E_i$ of the $ith$ ELM on the distribution $p(x)$ and the generalization error $\hat{E}$ of the ensemble on the distribution $p(x)$ are respectively:

$$E_i = \int dx p(x) E_i(x) \tag{26}$$

$$\hat{E} = \int dx p(x) \hat{E}(x) \tag{27}$$

We define the correlation between the $ith$ ELM and the $jth$ as:

$$C_{ij} = \int dx p(x)(f_i(x) - y)(f_j(x) - y) \tag{28}$$

Apparently, $C_{ij}$ satisfies equations (29) (30):

$$C_{ii} = E_i \tag{29}$$

$$C_{ij} = C_{ji} \tag{30}$$

Based on equations (23) (25), we have:

$$\hat{E}(x) = \left( \sum_{i=1}^{N} \omega_i f_i(x) - y \right)\left( \sum_{j=1}^{N} \omega_j f_j(x) - y \right) \tag{31}$$

Then consider equations (27), (28) and (31):

$$\hat{E} = \sum_{i=1}^{N} \sum_{j=1}^{N} \omega_i \omega_j C_{ij} \tag{32}$$

Since the optimal weight vector should minimize the generalization error $\hat{E}$ of the ensemble, the optimal weight vector $w_{opt}$ can be represented as:

$$w_{opt} = \arg\min_{\omega} \hat{E} = \arg\min_{w} \left( \sum_{i=1}^{N} \sum_{j=1}^{N} \omega_i \omega_j C_{ij} \right) \tag{33}$$

Let $w_{opt.k}$ be the $kth$ ($k = 1, 2, ..., N$) element in the optimal weight vector. $w_{opt}$ can be solved by Lagrange multiplier [37], which satisfies:

$$\frac{\partial \left( \sum_{i=1}^{N} \sum_{j=1}^{N} \omega_i \omega_j C_{ij} - 2\lambda \left( \sum_{i=1}^{N} \omega_i - 1 \right) \right)}{\partial w_{opt.k}} = 0 \tag{34}$$

Equation (34) can be simplified into:

$$\sum_{\substack{j=1 \\ j \neq k}}^{N} \omega_{opt.k} C_{kj} = \lambda \tag{35}$$

Since $w_{opt.k}$ satisfies equation (20), we can get:

$$\omega_{opt.k} = \frac{\sum_{j=1}^{N} C_{kj}^{-1}}{\sum_{i=1}^{N} \sum_{j=1}^{N} C_{ij}^{-1}} \tag{36}$$

Although equation (36) is enough to solve $w_{opt}$ in theory, it does not work well in practical applications because in the ensemble there are often several networks that are quite similar in performance, which leads the correlation matrix $(C_{ij})_{N \times N}$ of the ensemble to be an irreversible or ill-conditioned matrix (36). In order to estimate the fitness of the individuals in evolving population, we can use a validation dataset sampled from the training set. We can denote $\hat{E}_w^V$ to indicate the estimated generalization error of the ensemble corresponding to the individual $w$ on the validation data set $V$. Individual ELMs with larger weights than the threshold are selected to compose an optimal ensemble. Assuming that $N$ samples are generated from the training set, and each ELM is trained from training data, the ensemble output comes from the average output of the selective model. So finally, we can obtain the optimal ensemble ELM models using LARSEN-ELM for blended data. Above all, the pseudo code of LARSEN-ELM is organized as follows:

**Algorithm 4. LARSEN-ELM:** Input: Training set $\{X,Y\}$ (blended data) learner ELM, trials $N$, threshold $\lambda$,

*Step* 1: Preprocess $\{X,Y\}$ using LARS to form $\{X',Y'\}$.

*Step* 2: For $K=1$ to $N$ {
 $\{X_k, Y_k\}$ = samples from $\{X',Y'\}$
 Train individual ELM network by $\{X_k, Y_k\}$
 Calculate the predicted output $Y_{pre}$ }

*Step* 3: Generate a population of weight vectors

*Step* 4: Use selective ensemble to find the best weight vector $w_{opt}$.

*Step* 5: Output: $Y = Ave(\sum w_{opt} > \lambda * Y_{pre})$.

## 4. Experiments

In this section, we present some experiments to verify whether LARSEN-ELM performs better on robustness than ELM and other popular approaches such as OP-ELM, GASEN-BP, GASEN-ELM, and a kind of boosting algorithm called Least Squares Boosting(LSBoost) while keeping a high speed for blended data.

*4.1 a sum of two sines with one irrelevant noise variable*

The first case is a sum of two sines with one irrelevant noise variable, we provide 3 different irrelevant noise variables that all conform Gaussian distribution, such as $N(0,2)$, $N(0,1)$, $N(0,0.5)$. A set of 2001 points are generated as original training data, also a set of 20001 points as original testing data. Initially, we blend the noise variable with raw data(training and testing data), and then using blended training data, we apply ELM and LARSEN-ELM to train networks separately and get the regression output by blended testing data as shown below.

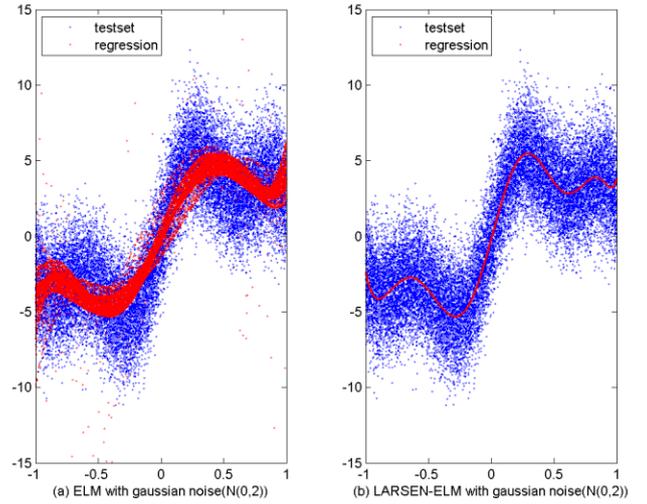

Figure 3. Results with Gaussian noise $N(0,2)$

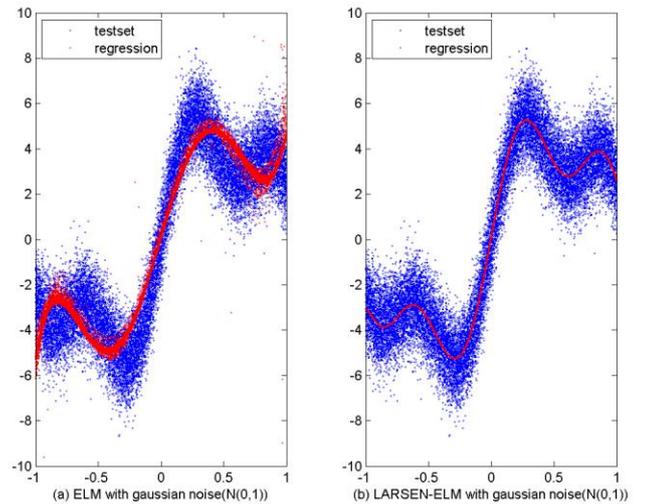

Figure 4. Results with Gaussian noise $N(0,1)$

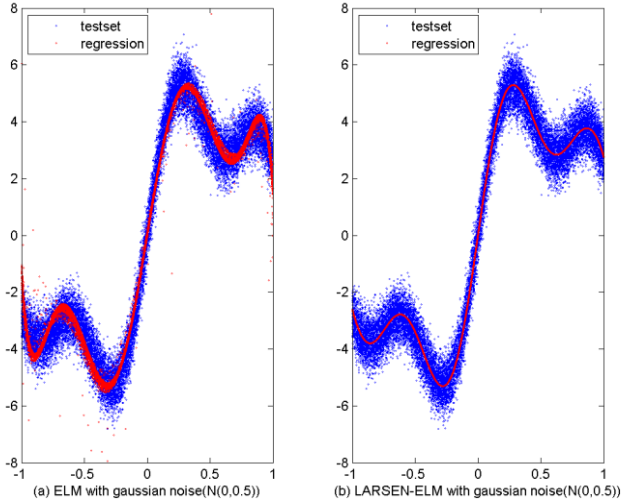

Figure 5. Results with Gaussian noise $N(0,0.5)$

Figure 3, Figure 4, and Figure 5 plot qualitatively the contrastive performance of ELM and LARSEN-ELM on robustness. In Figure 3, it shows that the Gaussian noise highly disturbs the robustness of ELM but has little effects on LARSEN-ELM. What's more, we can easily observe that the regression result deviates the tendency of the testing data in the ELM model, but the regression result keeps pace with the tendency of testing data in LARSEN-ELM model. The same things also occur in Figure 4 and Figure 5. Three comparative experiments commonly reflect that the robustness of LARSEN-ELM is better than the traditional ELM.

*4.2 four datasets from UCI machine learning repository*

The second case consists of four types of datasets from the UCI machine learning repository. The first one is Boston Housing dataset which contains 506 samples, including 13 input variables, and 1 output, this dataset is divided into a training set of 400 samples and the rest for testing set. The second one is Abalone dataset, there are 7 continuous input variables and 1 discrete input variable and 1 categorical attribute, and also it consists of 2000 training examples and 2177 testing examples in our experiments. The third one is Red Wine dataset which contains 1599 samples, 11 input variables, and 1 output, the dataset is divided into a training set of 1065 samples and the rest for testing set. Finally, for fully verifying our algorithm, Waveform dataset is selected because the number of input variables in waveform dataset (21 input variables) is comparatively higher than the datasets pervious provided.

Table 1
Specification of the four types of datasets

| Task | # variables | # training | # testing |
|---|---|---|---|
| Boston Housing | 13 | 400 | 106 |
| Abalone | 8 | 2000 | 2177 |
| Red Wine | 11 | 1065 | 534 |
| Waveform | 21 | 3000 | 2000 |

Firstly, we randomly blend several irrelevant Gaussian variables with the raw input variables, then all the features of data are preprocessed into a similar scale. Secondly, we compare our LASEN-ELM against ELM, OP-ELM, GASEN-BP, GASEN-ELM, and LSBoost in blended datasets. In our experiments, the genetic algorithm employed by LASEN-ELM is implemented by the GAOT toolbox developed by Houck et al. The genetic operators including selection, crossover probability, mutation probability, and stopping criterion, are all set to the default values of GAOT. The first group of original data is blended with 7 irrelevant variables that all conform to the Gaussian distributions, such as $N(0,2)$, $N(0,1)$, $N(0,0.5)$, $N(0,0.1)$, $N(0,0.005)$, $N(0,0.001)$, $N(0,0.0005)$. To make the experiment convincing, the second group of original data is blended with 10 irrelevant Gaussian variables, such as $N(0,2)$, $N(0,1)$, $N(0,0.5)$, $N(0,0.1)$, $N(0,0.05)$, $N(0,0.01)$, $N(0,0.005)$, $N(0,0.001)$, $N(0,0.0005)$, $N(0,0.0001)$. The threshold $\lambda$ used by GASEN is set to 0.05, and the initial number of ensemble components is set to 20 because we refer to some authoritative papers like *Generating Accurate and*

*Diverse Members of a Neural-Network Ensemble* (NIPS conference paper) [38] and Zhou's paper (GASEN). In addition, if the number of hidden units is set to 50, we can get a better performance because in this point (50) the testing RMSE curve becomes flat and the learning time is less [39]. For each algorithm, we perform 5 runs and record the average mean squared error and the computation time as well. The experimental results are tabulated in the following tables.

Table 2

Mean Square Error (MSE) for the 4 UCI datasets (7 irrelevant variables)

| Data set | ELM | OP-ELM | GASEN-BP | GASEN-ELM | LARSEN-ELM | LSBoost |
|---|---|---|---|---|---|---|
| Boston Housing | 33.9213 | 33.5064 | 26.7843 | 29.6033 | 20.8747 | 23.9058 |
| Abalone | 5.8258 | 5.1276 | 4.4499 | 4.9694 | 4.4723 | 4.7395 |
| Red Wine | 0.5092 | 0.4892 | 0.5277 | 0.4610 | 0.4407 | 0.5342 |
| Waveform | 0.3859 | 0.3460 | 0.2492 | 0.3415 | 0.2884 | 0.3126 |

Table 3

Computation Time (seconds) for the 4 UCI datasets (7 irrelevant variables, units: seconds)

| Data set | ELM | OP-ELM | GASEN-BP | GASEN-ELM | LARSEN-ELM | LSBoost |
|---|---|---|---|---|---|---|
| Boston Housing | 0.0219 | 0.0938 | 102.9344 | 0.8219 | 0.7156 | 0.5313 |
| Abalone | 0.0781 | 1.6406 | 323.4219 | 1.6781 | 1.5531 | 0.9938 |
| Red Wine | 0.0250 | 0.2563 | 230.9625 | 1.0469 | 1.1031 | 0.6469 |
| Waveform | 0.0750 | 1.6719 | 798.5875 | 1.7875 | 1.7750 | 1.5688 |

Table 4

Mean Square Error (MSE) for the 4 UCI datasets (10 irrelevant variables)

| Data set | ELM | OP-ELM | GASEN-BP | GASEN-ELM | LARSEN-ELM | LSBoost |
|---|---|---|---|---|---|---|
| Boston Housing | 33.5199 | 29.2905 | 27.5110 | 27.9480 | 22.3529 | 26.6808 |
| Abalone | 6.3571 | 5.1661 | 4.4580 | 5.8050 | 4.5057 | 4.7786 |
| Red Wine | 0.5011 | 0.5004 | 0.4986 | 0.4550 | 0.4506 | 0.5432 |
| Waveform | 0.3802 | 0.3281 | 0.2601 | 0.3480 | 0.2715 | 0.3175 |

Table 5

Computation Time (seconds) for the 4 UCI datasets (10 irrelevant variables, units: seconds)

| Data set | ELM | OP-ELM | GASEN-BP | GASEN-ELM | LARSEN-ELM | LSBoost |
|---|---|---|---|---|---|---|
| Boston Housing | 0.0156 | 0.1031 | 125.6219 | 0.7625 | 0.7063 | 0.5438 |
| Abalone | 0.0750 | 1.6281 | 413.1594 | 1.6406 | 1.5406 | 1.0938 |
| Red Wine | 0.0281 | 0.3063 | 257.1688 | 1.0156 | 0.9750 | 0.6750 |
| Waveform | 0.0656 | 1.6594 | 880.2188 | 1.8188 | 1.7750 | 1.6719 |

Table 6

Major Performance (MSE) comparison between new method and others (7 irrelevant variables) ( ">" means that new is better , "<" means that another is better)

| Data set | New & ELM | New & OP-ELM | New & GASEN-BP | New & GASEN-ELM | New & LSBoost |
|---|---|---|---|---|---|
| Boston Housing | > 38.46% | > 37.70% | > 22.06% | > 29.49% | > 12.68% |
| Abalone | > 23.23% | > 12.78% | < 0.50% | > 10.00% | > 5.64% |
| Red Wine | > 13.45% | > 9.91% | > 16.49% | > 4.40% | >17.50% |
| Waveform | > 25.27% | > 16.65% | 13.59% < | > 15.56% | > 7.74% |

Table 7

Minor Performance (Time) comparison between new method and others (7 irrelevant variables) ( ">" means that new is faster , "<" means that another is faster, units: seconds)

| Data set | New & ELM | New & OP-ELM | New & GASEN-BP | New & GASEN-ELM | New & LSBoost |
|---|---|---|---|---|---|
| Boston Housing | 0.6937 < | 0.6218 < | > 102.2188 | > 0.1063 | 0.1843 < |
| Abalone | 1.4750 < | > 0.0875 | > 321.8688 | > 0.1250 | 0.5593 < |
| Red Wine | 1.0781< | 0.8468 < | > 229.8594 | 0.0562 < | 0.4562 < |
| Waveform | 1.7000 < | 0.1031 < | > 796.8125 | > 0.0125 | 0.2062 < |

Table 8

Major Performance (MSE) comparison between new method and others (10 irrelevant variables) ( ">" means that new is better , "<" means that another is better)

| Data set | New & ELM | New & OP-ELM | New & GASEN-BP | New & GASEN-ELM | New & LSBoost |
|---|---|---|---|---|---|
| Boston Housing | > 33.31% | > 23.69% | > 18.75% | > 20.02% | > 16.22% |
| Abalone | > 29.12% | > 12.78% | < 1.06% | > 22.38% | > 5.71% |
| Red Wine | >10.08% | > 9.95% | > 9.63% | > 0.97% | > 17.05% |
| Waveform | > 28.59% | > 17.25% | 4.20% < | > 21.98% | > 14.49% |

Table9

Minor Performance (Time) comparison between new method and others (10 irrelevant variables) ( ">" means that new is faster , "<" means that another is faster, units: seconds)

| Data set | New & ELM | New & OP-ELM | New & GASEN-BP | New & GASEN-ELM | New & LSBoost |
|---|---|---|---|---|---|
| Boston Housing | 0.6907 < | 0.6218 < | > 124.9156 | > 0.0562 | 0.1625 < |
| Abalone | 1.4656 < | > 0.0875 | > 411.6188 | > 0.1000 | 0.4468 < |
| Red Wine | 0.9469 < | 0.6687 < | > 256.1938 | > 0.0406 | 0.3000 < |
| Waveform | 1.7094 < | 0.1156 < | > 878.4438 | > 0.0438 | 0.1031 < |

From above tables, we show that the percentage of MSE (major performance) drops using our method compared with others, and also show the comparison in running time (minor performance) among these algorithms. It can be concluded that the major performance of LARSEN-ELM is significantly better than that of ELM for blended data but their minor performance has a small gap, which means that our method is better than ELM in robustness with relatively high speed. We believe the selective ensemble and LARS selection contribute directly to the result; It also indicates that our method is better than OP-ELM in robustness because of selective ensemble technique. Although the new method is slightly worse than GASEN-BP in robustness, it saves a large amount of time. We believe ELM speeds up the training process; The new method performs better than GASEN-ELM because LARS definitely enhances the proportion of relevant inputs; Last but not least, the new method is obviously better than LSBoost which represents the classical ensemble methods on robustness with relatively high speed. From the above analysis, we believe that LARSEN-ELM can balance the robustness and speed effectively.

The following two observations are interesting.

Firstly, for Boston housing data with 7 irrelevant variables, the rate between the relevant and irrelevant variables is 13:7, and for Abalone data, the rate is 8:7, after LARS selection, for the Boston housing case, all selected input variables are from original dataset; For the Abalone case, 66% of selected input variables comes from original dataset, also the percentage of selected input variables coming from original dataset increases for both red wine dataset and waveform dataset. We find that LARS selection effectively improves the correlation between input variables and output variables, which largely benefit the robustness performance for blended data. The similar effects occur in the blended data with 10 irrelevant variables. Secondly, in the Abalone and Waveform dataset with irrelevant variables, the GASEN-BP is better than LARSEN-ELM at the cost of amounts of time, which proves that there are not any algorithms obtaining either the good robustness or high speed at the same time.

In general, we believe that our method is superior to the GASEN-BP because we have verified the goodness of LARSEN-ELM from different aspects. However, an important theoretical question is characterizing how different irrelevant variables affect robustness performance. We leave this question for further exploration.

## 5. Discussions

We applied our algorithm into three main experiments. In the first experiment, we added 1-dimensional Gaussian variable into a sum of two sines data. In the second experiment, we randomly added 7-dimensional Gaussian variables into four types of UCI datasets, so did the situation of 10 dimensions in the third experiment. All the results proved that the performance of our method on robustness is superior to the ELM and other approaches like LSBoost for blended data.

We select Gaussian noise as irrelevant variables because of its common in the real world. In LARSEN-ELM, the blended data is effectively preprocessed by LARS selection, so the input of preprocessed data is more related to the output of it. Moreover, the selective ensemble employs the genetic algorithm to select the optimal subset and ensembles them, so the robustness of network based on the optimal subset is definitely improved. Of course, our method also takes advantage of high speed in ELM.

We believe, for blended data, the ELM has a weak performance on robustness. Although the GASEN-BP has improved this situation, the training time is so long that we can't apply it into the real-time system. In contrast, our method has a good performance on robustness while keeping fast speed, so it has a great potential on real-time learning issues.

## 6. Conclusions

The proposed methodology (LARSEN-ELM) based on the extreme learning machine performs better than the original ELM for blended data, because of variables selection performed by LARS selection and selective ensemble. Variables selection not only ensures our method work well but also improves the ratio of original input variables after preprocessing. Moreover, when compared with GASEN-BP, in our method, though selective ensemble is so time-consuming, the computational time of our method is largely reduced because we employ ELM as basic units. Further research will investigate other combinations between LARS and the extensions of selective ensemble for performance enhancement such as eLARSEN-ELM which evolves from the extension of selective ensemble idea (Zhou et al.) etc. Recently, we will investigate a new method: train several LARSEN-ELMs at first, then collect them to constitute a new network by some specific rule. It may be an interesting work to develop a hierarchical method which can effectively increase performance in the future.

**Acknowledgments**

This work is partially supported by the High Technology Research and Development Program of China (2006AA09Z231),Natural Science Foundation of China(41176076, 31202036,51075377),and the Sci

## References

[1] G.-B. Huang, Q.-Y. Zhu, C.-K. Siew, Extreme learning machine: theory and applications, Neurocomputing, 70 (2006) 489-501.

[2] J. Ilonen, J.-K. Kamarainen, J. Lampinen, Differential evolution training algorithm for feed-forward neural networks, Neural processing letters, 17 (2003) 93-105.

[3] G. Bebis, M. Georgiopoulos, Feed-forward neural networks, IEEE Potentials, 13 (1994) 27-31.

[4] D.-E. Rumelhart, G.-E. Hinton, R.-J. Willians, Learning representations by back-propagating errors, Cognitive modeling, 1 (2002) 213.

[5] W. Jin, J.-L. Zhao, L.-S. Wei, et al., The improvements of BP neural network learning algorithm, WCCC-ICSP 2000. 5th International Conference on Signal Processing Proceedings, 3 (2000) 1647-1649.

[6] G.-B. Huang, D.-H. Wang, Y. Lan, Extreme learning machines: a survey, International Journal of Machine Learning and Cybernetics, 2 (2011) 107-122.

[7] G.-B. Huang, Q.-Y. Zhu, C.-K. Siew, Extreme learning machine: a new learning scheme of feedforward neural networks, IEEE International Joint Conference on Neural Networks, 2 (2004) 985-990.

[8] J.-P. Deng, N. Sundararajan, P. Saratchandran, Communication channel equalization using complex-valued minimal radial basis function neural networks, IEEE Transactions on Neural Networks, 13 (2002) 687-696. [9] G. Li, M. Liu, M. Dong, A new online learning algorithm for structure-adjustable extreme learning machine, Computers & Mathematics with Applications, 60 (2010) 377-389.

[10] G.-B. Huang, L. Chen, Convex incremental extreme learning machine, Neurocomputing, 70 (2007) 3056-3062.

[11] B. Frénay, M. Verleysen, Parameter-insensitive kernel in extreme learning for non-linear support vector regression, Neurocomputing, 74 (2011) 2526-2531.

[12] G.-B. Huang, H. A. Babri, Upper bounds on the number of hidden neurons in feedforward networks with arbitrary bounded nonlinear activation functions, IEEE Transactions on Neural Networks, 9 (1998) 224-229.

[13] Y. Miche, A. Sorjamaa, P. Bas, et al., OP-ELM: optimally pruned extreme learning machine, IEEE Transactions on Neural Networks, 21 (2010) 158-162.

[14] Y. Miche, A. Sorjamaa, A. Lendasse, OP-ELM: theory, experiments and a toolbox, Artificial Neural Networks-ICANN 2008, ed: Springer, 2008, pp. 145-154.

[15] Y. Miche, P. Bas, C. Jutten, et al., A methodology for building regression models using extreme learning machine: OP-ELM, Proceedings of the European Symposium on Artificial Neural Networks (ESANN), (2008) 247-252.

[16] L. K. Hansen, P. Salamon, Neural network ensembles, IEEE Transactions on Pattern Analysis and Machine Intelligence, 12 (1990) 993-1001.

[17] Z.-L. Sun, T.-M. Choi, K.-F. Au, et al., Sales forecasting using extreme learning machine with applications in fashion retailing, Decision Support Systems, 46 (2008) 411-419.

[18] D. Slepian, The one-sided barrier problem for Gaussian noise, Bell System Tech. J, 41 (1962) 463-501.

[19] M. van Heeswijk, Y. Miche, T. Lindh-Knuutila, et al., Adaptive ensemble models of extreme learning machines for time series prediction, Artificial Neural Networks–ICANN 2009, ed: Springer, 2009, pp. 305-314.

[20] Y. Tang, B. Biondi, Least-squares migration/inversion of blended data, SEG Technical Program Expanded Abstracts, 2009, pp. 2859-2863.

[21] T. Similä, J. Tikka, Multiresponse sparse regression with application to multidimensional scaling, Artificial Neural Networks: Formal Models and Their Applications–ICANN 2005, ed: Springer, 2005, pp. 97-102.

[22] Z.-H. Zhou. J.-X. Wu, J. Yuan, et al., Genetic algorithm based selective neural network ensemble, IJCAI-01: Proceedings of the Seventeenth International Joint Conference on Artificial Intelligence, Seattle, Washington, August 4-10, 2001, pp. 797.

[23] N. Li, Z.-H. Zhou, Selective ensemble under regularization framework Multiple Classifier Systems, ed: Springer, 2009, pp. 293-303.

[24] A. Asuncion, D.-J. Newman, UCI machine learning repository, 2007.

[25] G.-B. Huang, C.-K. Siew, Extreme learning machine: RBF network case, Control, Automation, Robotics and Vision Conference, 2004. ICARCV 2004 8th, 2 (2004) 1029-1036.

[26] G.-B. Huang, Q.-Y. Zhu, C.-K. Siew, Extreme learning machine: a new learning scheme of feedforward neural networks, IEEE International Joint Conference on Neural Networks, 2004, 2 (2004) 985-990.

[27] G.-B. Huang, Learning capability and storage capacity of two-hidden-layer feedforward networks, IEEE Transactions on Neural Networks, 14 (2003) 274-281.

[28] C.-R. Rao, S.-K. Mitra, Generalized inverse of a matrix and its applications, J. Wiley, New York, 1972.

[29] D. Serre, Matrices: Theory and Applications. 2002, ed: Springer, New York, 2002.

[30] O. Axelsson, A generalized conjugate gradient, least square method, Numerische Mathematik, 51 (1987) 209-227.

[31] G. Chavent, Identification of distributed parameter systems: about the output least square method, its implementation and identifiability, Proc. 5th IFAC Symposium on Identification and System Parameter Estimation, (1979) 85-97.

[32] L. Breiman, "Bagging predictors," Machine learning, 24 (1996) 123-140.

[33] B. Atoufi and H.S. Hosseini, "Bio-Inspired Algorithms for Fuzzy Rule-Based Systems ", Advanced Knowledge Based Systems: Model, Applications and Research, 1(2010)126-159.

[34] S. Barai and Y. Reich, Ensemble modelling or selecting the best model: Many could be better than one, AI EDAM, 13 (1999) 377-386.

[35] L.-J. Zhao, T.-Y. Chai, D.-C. Yuan, Selective ensemble extreme learning machine modeling of effluent quality in wastewater treatment plants, International Journal of Automation and Computing 9 (2012) 627-633.

[36] J.-X. Wu, Z.-H. Zhou, and Z.-Q. Chen, et al., Ensemble of GA based selective neural network ensembles, Proceedings of the 8th International Conference on Neural Information Processing, Shanghai, China, 2001, 3 (2001) 1477-1483.

[37] R. Bellman, Dynamic programming and Lagrange multipliers, Proceedings of the National Academy of Sciences of the United States of America 42 (1956) 767.

[38] D.W. Opitz, J.W. Shavlik, Generating Accurate and diverse members of a neural-network ensemble, advances in neural information processing systems 8, MIT Press: Cambridge,MA.

[39] G.B. Huang, L. Chen and C.K. Siew,universal approximation using incremental constructive feedforward networks with random hidden nodes, EEE Transactions on Neural Networks, 17 (2006) 4.

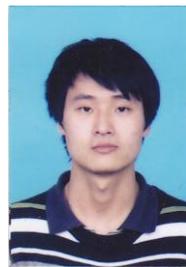
**Bo Han** is a M.S. student at the College of Information Science and Engineering, Ocean University of China. His major is signal and information processing, and he will receive M.S. degree from department of Electronics Engineering at College of Information Science and Engineering in 2014. Currently his primary research interests include machine learning, computer vision.

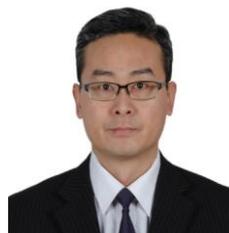
**Bo He** received his M.S. and Ph.D. degrees from Harbin Institute of Technology, China, in 1996 and 1999 respectively. From 2000 to 2003, Dr HE has been with Nanyang Technological University (Singapore) as a Post-Doctoral Fellow, where he worked on mobile robots, unmanned vehicles, research works included precise navigation, control and communication. In 2004, Dr HE joined Ocean Univ. of China (OUC), now he is a full professor of OUC and deputy head of department of Electronics Engineering at College of Information Science and Engineering. Currently his research interests include AUV design and applications, AUV SLAM, AUV control, and machine learning.

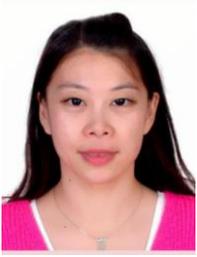
**Rui Nian** is an assistant professor in the Information Science and Engineering College at Ocean University of China (OUC) now. She received her B.S. and M.S. Degree in signal and information processing from OUC, and got her Doctor degree respectively from OUC and Université catholique de Louvain (UCL). Her primary research interests include computer vision, pattern recognition, image processing, cognitive science, statistical learning and high dimensional space theory. She is also jointly appointed as the visiting professor in the ICTEAM, UCL and the research scientist in the the Research and Development Center, Haier Group.

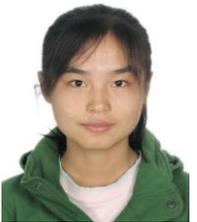
**Mengmeng Ma** is a M.S. student at the College of Information Science and Engineering, Ocean University of China. Her major is communication and information system. Currently her primary research interests include machine learning, computer vision.

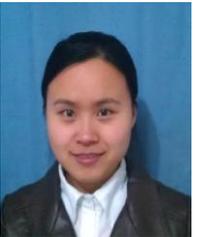
**Shujing Zhang** is a Ph.D. student at the College of Information Science and Engineering, Ocean University of China. Her major is ocean information detection and processing. Currently her research interests include SLAM for AUVs and machine learning.

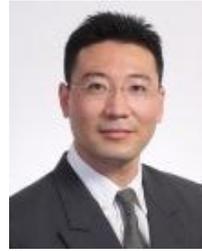
**Minghui Li** is a lecture in the Department of Electronic and Electrical Engineering at the University of Strathclyde. He received his Ph.D. degree in electrical engineering from Nanyang Technological University, Singapore in 2004 for work on array processing algorithms and systems. His research focuses on ultrasound system design and various signal processing problems faced by ultrasonic applications including coded excitation, ultrasound beamforming, array design and synthesis.
http://www.strath.ac.uk/eee/research/cue/staff/drdavidli/

minghui.li@strath.ac.uk

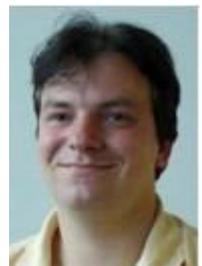
**Amaury Lendasse** is a Docent and senior researcher in ICS, HUT. He got his M.S. Degree in Mechanical Engineering in the Universite de Louvain-la-Neuve in Belgium in 1997 and a second M.S. Degree in Control in the same university in 1997. He received his Ph.D. degree in 2003 in Louvain-la-Neuve. His research interests lie in Machine Learning, Time-Series Prediction, Environmental Modeling, Industrial Applications, Information Security, Variable Selection, GPU.

Amaury.lendasse@aalto.fi